\documentclass[runningheads]{llncs}
\usepackage{graphicx}
\usepackage{amsmath}
\usepackage[utf8]{inputenc}
\begin{document}
\title{Unsupervised Temporal Video Segmentation as an Auxiliary Task for Predicting the Remaining Surgery Duration\thanks{Funded by the German Research Foundation (DFG, Deutsche 
Forschungsgemeinschaft) as part of Germany’s Excellence Strategy – 
EXC 2050/1 – Project ID 390696704 – Cluster of Excellence “Centre for 
Tactile Internet with Human-in-the-Loop” (CeTI) of Technische 
Universität Dresden.}}
\titlerunning{Unsupervised Temporal Video Segmentation for RSD Prediction}
%
\author{Dominik Rivoir\inst{1,3} \and
Sebastian Bodenstedt\inst{1} \and
Felix von Bechtolsheim\inst{2} \and
Marius Distler\inst{2} \and
Jürgen Weitz\inst{2,3} \and
Stefanie Speidel\inst{1,3}}
%

\authorrunning{D. Rivoir et al.}
%
\institute{Translational Surgical Oncology, National Center for Tumor Diseases (NCT), Dresden, Germany\\
\email{\{firstname.lastname\}@nct-dresden.de} \and
Department of Visceral, Thoracic and Vascular Surgery, Faculty of Medicine
and University Hospital Carl Gustav Carus, Technische Universität Dresden, Germany \and
Centre for Tactile Internet with Human-in-the-Loop (CeTI), TU Dresden, Germany}
\maketitle              
\begin{abstract}
Estimating the remaining surgery duration (RSD) during surgical procedures can be useful for OR planning and anesthesia dose estimation. With the recent success of deep learning-based methods in computer vision, several neural network approaches have been proposed for fully automatic RSD prediction based solely on visual data from the endoscopic camera. We investigate whether RSD prediction can be improved using unsupervised temporal video segmentation as an auxiliary learning task. As opposed to previous work, which presented supervised surgical phase recognition as auxiliary task, we avoid the need for manual annotations by proposing a similar but unsupervised learning objective which clusters video sequences into temporally coherent segments. In multiple experimental setups, results obtained by learning the auxiliary task are incorporated into a deep RSD model through feature extraction, pretraining or regularization. Further, we propose a novel loss function for RSD training which attempts to counteract unfavorable characteristics of the RSD ground truth. Using our unsupervised method as an auxiliary task for RSD training, we outperform other self-supervised methods and are comparable to the supervised state-of-the-art. Combined with the novel RSD loss, we slightly outperform the supervised approach.

\end{abstract}
\section{Introduction}
Resources in the operating room (OR) are among the most expensive in a hospital and careful OR planning is crucial in order to minimize waiting times and idle phases. Estimating the remaining surgery duration (RSD) at specified points during an intervention can facilitate more efficient utilization of OR resources.\\
This work builds on deep learning-based methods for fully automated RSD prediction based solely on endoscopic video data~\cite{rsd_w_phase,rsd_w_instruments,rsd_net}. Since the remaining time for each frame can be inferred automatically from a given videos, RSD prediction is a self-supervised task. This property is especially useful in medical applications, where manually annotating data is expensive.\\
However, RSD prediction is an extremely challenging task due to the complexity and uniqueness of a surgical procedure. It appears to require a high-level understanding of the workflow and progress of the surgery. These factors probably contribute to RSD models tending to overfit without proper regularization or pretraining~\cite{rsd_net}. To alleviate this problem, Twinanda et. al. propose an RSD prediction network which is encouraged to learn progress-related features and utilizes the elapsed time in addition to visual features~\cite{rsd_net}. Bodenstedt et. al. use multimodal sensor data from the OR including visual data and tool signals for their prediction~\cite{rsd_w_instruments}. State-of-the-art results are obtained by Aksamentov et. al. who suggest to pretrain the RSD model on surgical phase recognition as an auxiliary task~\cite{rsd_w_phase}. However, surgical phase recognition is a supervised task and therefore reduces the advantages of self-supervised RSD training.\\
Our contributions consist of proposing an unsupervised auxiliary task to improve RSD prediction, namely unsupervised temporal video segmentation. To solve the auxiliary task, we present a method for finding segmentations that capture the progress of a surgery similar to surgical phases but without the need for manual annotations. As indicated in \cite{rsd_w_phase,rsd_net}, progress-related features can be beneficial for RSD prediction. Using an unsupervised auxiliary task makes this approach widely applicable to different datasets. Several image-based unsupervised temporal video segmentation methods have been proposed~\cite{kukleva,yao,spr_unsup}. We adopt the method from \cite{yao} since its iterative procedure allows us to learn task-related image features. The other approaches extract or learn features prior to segmentation, making them unsuitable as an auxiliary task. Finally, we propose a novel loss function that targets undesirable characteristics of the RSD ground truth.

\section{Methods}
Our approach combines models for RSD prediction and unsupervised temporal video segmentation. A model consisting of a Convolutional Neural Network (CNN) for visual feature extraction and a Long Short-Term Memory network (LSTM) for propagating information through time is trained to perform our main task, RSD prediction, similar to \cite{rsd_w_phase,rsd_w_instruments,rsd_net}. For the temporal segmentation task, we use an unsupervised approach to train a discriminative-generative model alternating between learning segmentation labels through a generative model and learning visual features in a discriminative CNN-LSTM network. The results obtained by solving the temporal segmentation task can be leveraged for RSD prediction in several ways. First, we assume that the temporal segmentation training encourages the CNN-LSTM model to learn features relevant for RSD prediction. Thus, we investigate reusing the learned feature representations by initializing the CNN-LSTM model for RSD prediction with the learned network weights. We then pursue two different strategies for further training the RSD model: we either finetune only the upper layers or none of the layers in the CNN. In a complementary approach, we use the obtained segment labels to formulate an additional objective to regularize the RSD model during training.
\begin{figure}[t]
\centering
\includegraphics[width=.96\linewidth]{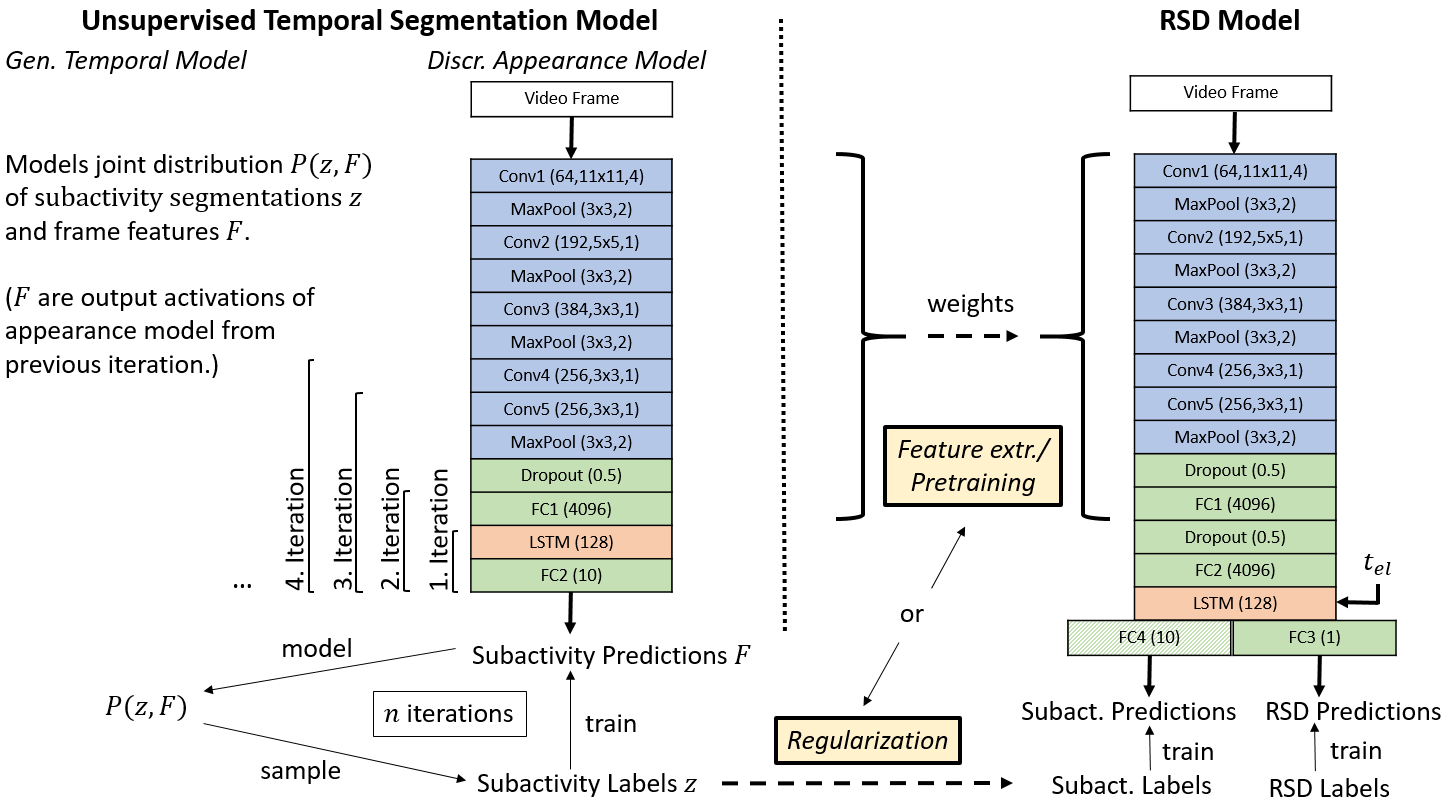}
\caption{Summary of the proposed learning pipelines. Step 1: Train the unsupervised temporal segmentation model for $n$ iterations. Step 2: Either use the learned weights for feature extraction or pretraining or the learned segmentation labels for regularization. Note that \emph{FC4} only exists in the regularization pipeline. Network layer notation: convolutional layers \emph{Conv* (filter size, kernel size, stride)}, max-pooling layers \emph{MaxPool (kernel size, stride)}, dropout layers \emph{Dropout (drop probability)} and fully-connected and recurrent layers \emph{FC*/LSTM (size)}.}
\label{fig:pipeline}
\end{figure}

\subsection{RSD Model} \label{sec:rsd_model}
For our RSD model (Fig. \ref{fig:pipeline}, right), we use an AlexNet-style CNN~\cite{cnn} to extract visual features from the video frames of a recorded surgical procedure. The feature representations are concatenated with the elapsed time $t_{el}$ of the procedure and fed into an LSTM, similar to \cite{rsd_net}. The LSTM can consider features from the current and previous frames and produces an RSD estimate for each frame of the video. The network predicts the remaining duration in minutes, scaled by a factor of $0.05$ due to high values of up to 100 minutes. RSD prediction is formulated as a regression task and optimized according to the SmoothL1 loss~\cite{rsd_net}. We use a simpler model instead of \emph{RSDNet}~\cite{rsd_net}, since the latter showed no empirical improvement in combination with our auxiliary task.

\subsection{Unsupervised Temporal Video Segmentation Model} \label{sec:utvs}
\begin{figure}[t]
\centering
\begin{minipage}{.6\textwidth}
	\centering
	\includegraphics[width=\linewidth]{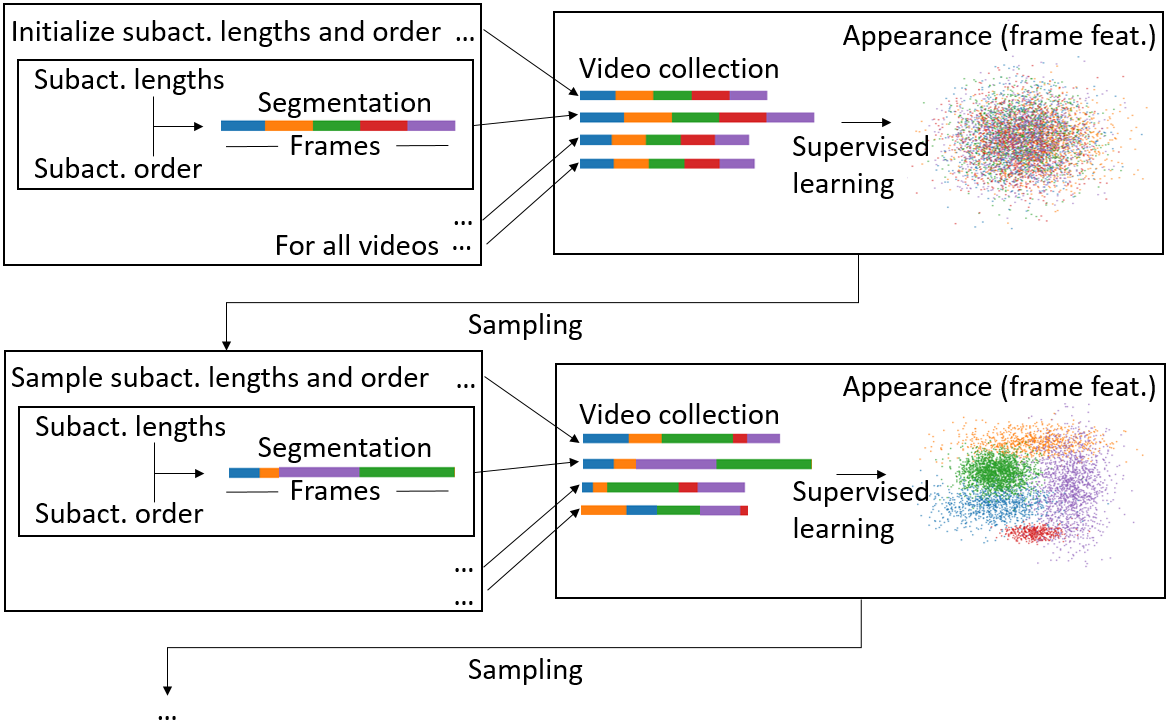}
	\caption{The unsupervised temporal segmentation method adopted from \cite{yao}. We alternate between learning frame features and subactivity labels.}
	\label{fig:activity}
\end{minipage}
\hfill
\begin{minipage}{.36\textwidth}
	\centering
	\includegraphics[width=\linewidth]{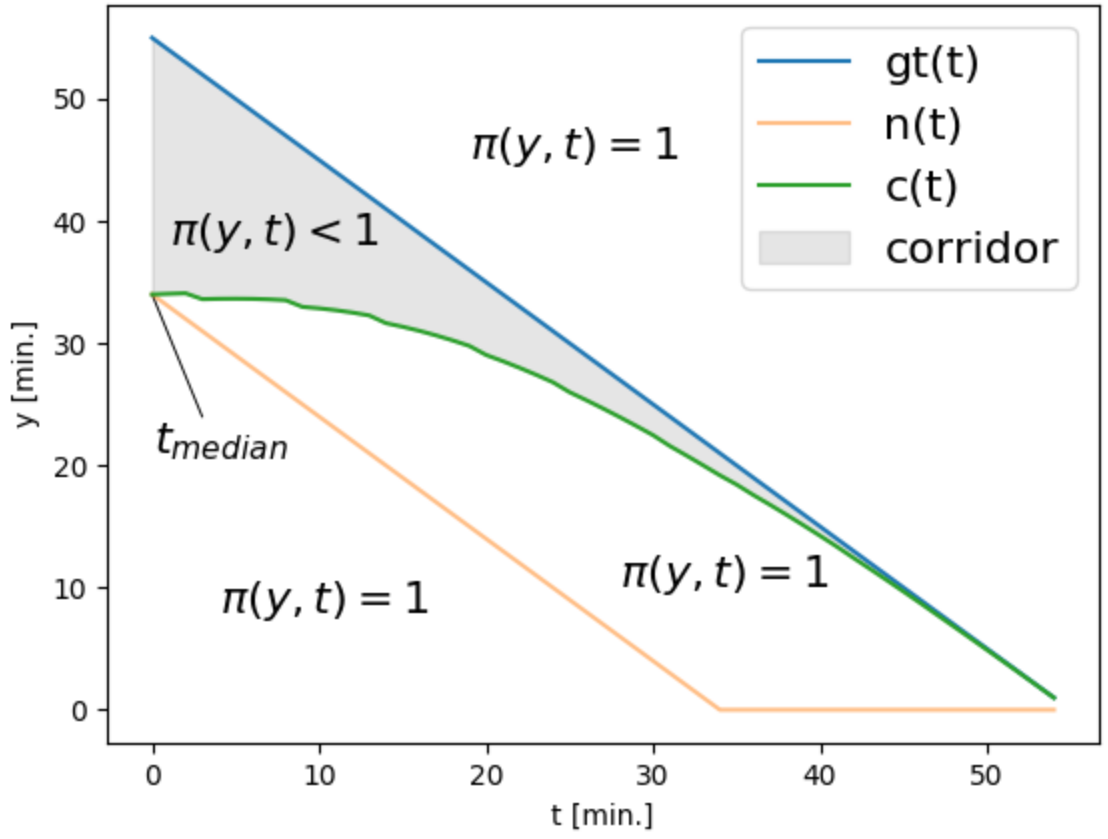}
	\caption{Ground truth $gt(t)$, corridor border $c(t)$ and median-based prediction $n(t)$ for one surgery.}
	\label{fig:corridor}
\end{minipage}
\end{figure}

We extend a method from \cite{yao} for recognition and segmentation of complex activities in videos, i.e. activities consisting of several subactivities. The author's definiton of a complex activity can be applied to surgeries, where subactivities could represent surgical phases or similar steps.\\
The unsupervised learning algorithm alternates between learning frame features and subactivity labels (Fig. \ref{fig:activity}). Given the current subactivity labels, a discriminative appearance model learns frame features in a supervised manner. A generative temporal model is then estimated, which models the distribution of subactivity lengths and subactivity orders, given the distribution of frames in the learned appearance space. The subactivity lengths and order determine the segmentation of a video. After sampling new lengths and orders and subsequently updating subactivity labels, the algorithm continues to learn new frame features.\\
The discriminative appearance model is a CNN-LSTM model (Fig. \ref{fig:pipeline}, left) optimized via the cross-entropy loss. An extensive hyperparameter search suggested the use of ten subactivities. Opposed to our deep learning approach, the appearance model in the original paper \cite{yao} learns a simple linear embedding of image features. When replacing this simple model by a complex CNN-LSTM model, care must be taken to avoid overfitting on unrefined segmentations from early iterations. To this end, only the top two layers of the network are optimized in the first iteration and layers are added incrementally after each iteration (Fig. \ref{fig:pipeline}, left). In turn, the incremental depth increase requires an initialization of the fixed layers. We pretrain the CNN using the 2nd-order temporal coherence objective~\cite{higher_order}, which has shown promising results on a similar task~\cite{tc_spr}.\\
The generative temporal model estimates the joint distribution of frame features and subactivity segmentations. The distribution over segmentations is modeled by distributions over the length of each subactivity \emph{(Multinomial)} and over the order of subactivities \emph{(Generalized Mallows Model)}. Sampling-based approximations are used to infer segmentations. The generative temporal model is almost identical to the one proposed in \cite{yao}. We only drop background model.\\
The method produces new models after each iteration. Hence, we need to evaluate and select a model to use as a support for the RSD model. Since the ground truth segmentation labels are unknown, we require a surrogate quality measure. We define a measure $TC$ which quantifies the temporal coherence of subacitivty predictions by the appearance model. More precisely, we measure the prediction's accuracy with respect to the best match of coherent segmentations with the same subacitivity lengths. This measure intends to capture how well a model has learned progress-related features. Preliminary experiments showed that the measure selects models which are beneficial for RSD prediction.

\subsection{Combined Learning Pipelines} \label{sec:pipe}
Fig. \ref{fig:pipeline} shows three strategies for combining models.\\
\textbf{Feature extraction:} 
The unsupervised temporal segmentation method is used to train the CNN-LSTM network of the discriminative appearance model. The weights learned from layers \emph{Conv1} to \emph{FC1} are then re-used for the RSD model. While training the RSD model, the initialized layers are fixed. Only layers \emph{FC2}, \emph{LSTM} and \emph{FC3} are optimized. This method is equivalent to feature extraction, where layers \emph{Conv1} to \emph{FC1} serve as a feature extractor for a shallow RSD model.\\
\textbf{Pretraining:} 
Pretraining is almost identical to feature extraction, except that the layers \emph{Conv5} and \emph{FC1} are optimized during RSD training after being initialized by the temporal segmentation method. In order to prevent the previously learned information from being overwritten too quickly, a lower learning rate is applied to pretrained layers. To summarize, layers \emph{Conv1} to \emph{Conv4} are fixed, \emph{Conv5} and \emph{FC1} are optimized with a low learning rate, and \emph{FC2}, \emph{LSTM} and \emph{FC3} are optimized using the regular learning rate.\\
\textbf{Regularization:} 
The resulting subactivity labels of a learned temporal segmentation model are re-used for supervision during RSD training. First, segmentations are learned for each video by the unsupervised temporal segmentation model. Then, the RSD model is jointly trained on RSD prediction and predicting the current subactivity according to the previously found segmentations.
\subsection{Corridor-based RSD Loss Function}
In the early stages of a procedure it is extremely challenging to correctly predict the remaining duration since later occurring events are not yet known. To account for this, we propose an alternative RSD loss function which reduces the influence of early errors. Intuitively, we do not want to penalize the best guess at the beginning of a procedure, which is the average length of the given procedure type. For each video, we therefore define an area between the ground truth $gt(t)$ over time and a na\"ive median-based prediction $n(t) = max(t_{median} - t, 0)$, where $t_{median}$ is the median duration of all procedures in the training set. Errors within this corridor are decreased by a weighting function $\pi$ (Fig. \ref{fig:corridor}). The corridor border $c(t)=\alpha_t gt(t) + (1-\alpha_t) n(t)$ is a linear combination of the ground truth $gt(t)$ and the median-based prediction $n(t)$.\\
Here $\alpha_t= 1 - \frac{2}{1+e^{5\cdot prog(t)}}$ is a time-dependent linear factor similar to the tanh function, where $prog(t) = \frac{t}{gt(t)+t}$ is the progress of the surgery in percent. Intuitively, $c(t)$ is closer to the median-based prediction $n(t)$ at early time points, when little information is available, and approaches the ground truth $gt(t)$ as the procedure progresses. The weight $\pi(y,t)$ for a prediction $y$ at time $t$ is given by
\begin{equation}
\pi(y,t)=
\begin{cases}
\left(\frac{|y-gt(t)|}{|c(t)-gt(t)|}\right)^2, & \text{if } c(t) \leq y \leq gt(t) \text{ or } gt(t) \leq y \leq c(t)\\
1, & \text{otherwise}
\end{cases}
\end{equation}
$\pi$ realizes a smooth weighting distribution along the $y$-axis inside the corridor from $y=gt(t)$ to $y=c(t)$ (with $\pi(gt(t),t)=0$ to $\pi(c(t),t)=1$). For predictions $y$ outside the corridor, $\pi(y,t)=1$. The corridor-weighted loss is finally given by
\begin{equation}
CorrSmoothL1(y,t) = \pi(y,t) \cdot SmoothL1(y,gt(t))
\end{equation}

\section{Evaluation}
We evaluate our proposed models on the publicly available Cholec80 dataset~\cite{endonet}. We use 50 videos for training, 10 for validation and 20 for testing. Video frames are extracted at 1fps. We train the RSD models using the Adam optimizer (200 epochs, learning rate $10^{-5}$, batch size 384, $\mathcal{L}_2$-weight $10^{-5}$). For the pretraining pipelines, we use SGD, run 250 epochs and update pretrained layers with a learning rate of $10^{-6}$ since these settings empirically perform better. The other settings are kept. For the segmentation model, Adam, learning rate $10^{-5}$, batch size 384, 5 epochs per iteration, 8 iterations, $\mathcal{L}_2$-weight $10^{-4}$ are used. We select the best model from iterations 6 to 8 according to our TC measure (Sec. \ref{sec:utvs}).
\subsection{Baselines}
We consider four baselines for RSD prediction: The simplest baseline is the RSD model from Section \ref{sec:rsd_model} trained only on single-task RSD prediction with no auxiliary task \emph{(None)}. The other baselines are supported by auxiliary tasks each using all three proposed pipelines from Sec. \ref{sec:pipe}. The first auxiliary task is temporal segmentation into 10 uniform segments \emph{(Uniform)}. This is an interesting baseline that can provide insight into how much RSD-relevant information is gained by learning more refined segmentations. The other two auxiliary tasks are state-of-the-art approaches, namely supervised surgical phase recognition \emph{(Phase)}~\cite{rsd_w_phase} and self-supervised prediction of progress $prog(t)$ \emph{(Progress)} reimplemented from \cite{lessismore}, which is an updated version of the RSDNet from \cite{rsd_net}. For the phase approach, we use the regularized RSD model from Fig. \ref{fig:pipeline} in order preserve compariability to our methods. The main differences to the architecture from \cite{rsd_w_phase} are that we use an AlexNet-style CNN like in \cite{lessismore} and that we incorporate the elapsed time into the prediction like in \cite{rsd_net,lessismore}. Hyperparameters of the optimization are identical to the proposed methods.
\subsection{Results}
\begin{figure}[t]
\centering
\includegraphics[width=.94\linewidth]{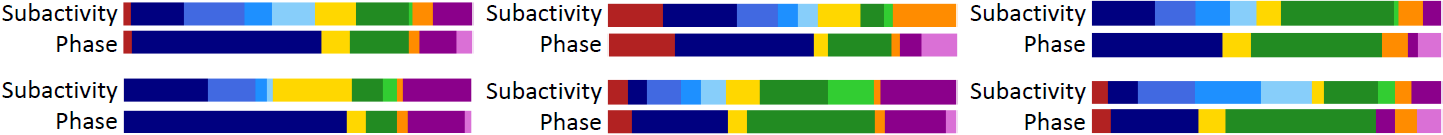}
\caption{Ground truth of surgical phases and learned subactivities of exemplary training videos. Different shades illustrate how several subactivities correlate with one phase or vice versa. Our method does not provide a mapping from subactivities to phases.}
\label{fig:segs}
\end{figure}

\begin{table}[t]
\centering
\resizebox{.85\linewidth}{!}{
\begin{tabular}{|l|c|c|c|}
\hline
\textbf{Auxiliary Task} & \textbf{Feature Extraction} & \textbf{Pretraining} & \textbf{Regularization}\\
\hline
Unsup. temp. seg. \textit{(ours)} & 9.0 $(\pm 0.1)$ & 9.3 $(\pm 0.2)$ & 9.2 $(\pm 0.5)$ \\
\hline
None & \multicolumn{3}{|c|}{9.7 $(\pm 0.1)$}\\
Uniform & 9.4 & 9.4 & 9.4\\
Progress & 9.0 $(\pm 0.1)$ & 9.5 & 9.6\\
Phase \textit{(supervised)} & \textbf{8.9} $(\pm 0.1)$ & \textbf{8.9} & 9.1\\
\hline
\end{tabular}}
\caption{Mean average error (MAE) in minutes for our proposed RSD models as well as state-of-the-art baselines.}
\label{fig:res1}
\end{table}
\begin{table}[t]
\centering
\resizebox{.9\linewidth}{!}{
\begin{tabular}{|l|c|c|c|c|}
\hline
& \multicolumn{2}{|c|}{\textbf{Feature extraction}} & \multicolumn{2}{|c|}{\textbf{Regularization}}\\
\hline
\textbf{Auxiliary Task} & \textbf{SmoothL1} & \textbf{CorrSmoothL1} & \textbf{SmoothL1} & \textbf{CorrSmoothL1}\\
\hline
Unsup. temp. seg. \textit{(ours)} & 9.0 $(\pm 0.1)$ & 9.1 $(\pm 0.2)$ & 9.2 $(\pm 0.5)$ & \textbf{8.7} $(\pm 0.2)$\\
\hline
None* & 9.7 $(\pm 0.1)$ & 9.1 $(\pm 0.5)$ & 9.7 $(\pm 0.1)$ & 9.1 $(\pm 0.5)$\\
Uniform & 9.4 & 9.3 & 9.4 & 9.4\\
Progress & 9.0 $(\pm 0.1)$ & 9.1 & 9.6 & 9.2 $(\pm 0.4)$\\
Phase \textit{(supervised)} & 8.9  $(\pm 0.1)$ & 9.0 & 9.1 & 8.9 $(\pm 0.1)$\\
\hline
\end{tabular}}
\caption{Comparison of RSD loss functions on the feature extraction and regularization pipelines. *In \emph{None}, columns 2 and 4 as well as 3 and 5 refer to the same experiments.}
\label{fig:res2}
\end{table}
Table \ref{fig:res1} shows the mean average error (MAE) in minutes for each of our proposed models as well as all baselines using the SmoothL1 loss. All experiments involving our proposed method are performed four times, averaged and indicated by a standard deviation. Baseline experiments for settings which were effective for our method are repeated four times, in order to obtain more significant results.\\
Comparing our proposed methods, feature extraction achieves the best results ($9.0\pm0.1$ min. MAE), while pretraining performs worst ($9.3\pm0.2$) and high variances were observed during regularization ($9.2\pm0.5$). The high expressivity of RSD models likely causes overfitting in the two latter setups. In the pretraining setting, the RSD model is the least expressive, as only the top layers are optimized after initialization by the segmentation method. Hence, it is supposedly less prone to overfitting. We also observe that our approach outperforms or matches the self-supervised approaches (single-task RSD, uniform segmentation and progress) for all learning pipelines. Using feature extraction, we even achieve results comparable to the supervised phase-based approach (9.0 vs. 8.9).\\
Next, we compare the CorrSmoothL1 loss to SmoothL1 on the previously most successful feature extraction and the regularization pipeline (Table \ref{fig:res2}), since the high variance in regularization experiments indicates potential for improvement. The first two result columns show RSD errors for both loss functions on the feature extraction pipeline. No clear difference can be observed. The single-task RSD model as well as most regularized models, however, improve drastically. Since CorrSmoothL1 aims to reduce overfitting, it is more effective on very expressive deep models such as the regularization models or the single-task model. In the feature extraction setup, which has significantly fewer trained parameters during RSD training, the model's low expressivity probably prevents further improvement. Using regularization, our approach improves from 9.2 to 8.7 minutes MAE and therefore exceeds our previously best result as well as all baselines. We even outperform all supervised phase-based setups. It is not clear how significant this difference is, since the supervised approach performed slightly better than ours in the SmoothL1 setup. However, even comparable results are very promising and our approach performs at least on a similar level as supervised methods. Fig. \ref{fig:segs} shows that subactivity labels corresponds fairly well to surgical phases but are more fine grained due to the higher number of segments. Using a hand-picked mapping from subactivities to phases, we achieve an accuracy of 71\% and 72\% on the training and test set for surgical phase recognition. A limitation of our proposed method remains the complexity of the whole model and achieving stable results poses a challenge.
\section{Conclusion}
We present unsupervised temporal video segmentation as a novel auxiliary task for video-based RSD prediction and propose three different learning pipelines to utilize unsupervised temporal segmentation learning for RSD modeling. In our experiments on the Cholec80 dataset, our approach compares favorably with self-supervised auxiliary tasks and performs comparably to the state of the art, which utilizes supervised surgical phase recognition as auxiliary task. This is very promising since the method does not require any manual annotations and therefore has potential for improvement by utilizing larger, unlabeled datasets. Further, we specifically target the problem that RSD ground truth labels can be misleading in early stages of a procedure. Our novel corridor-based loss shows clear improvements on deep RSD models. Using the corridor-based loss, we even outperform the state of the art when we regularize the RSD model with the unsupervised temporal segmentation task. Future work could evaluate our method on procedure types with higher variance in duration and therefore lower correlation between RSD and progress. Analyzing how our method transfers to these procedures is interesting since temporal segmentations can potenially model more complex temporal structures than progress. Also, the similarity of unsupervised segmentations and surgical phases induces interesting new research directions.

%
%
%

\end{document}